\documentclass[conference]{IEEEtran}
\usepackage{amsmath,amssymb,amsfonts}
\usepackage{algorithmic}
\usepackage{graphicx}
\usepackage{textcomp}
\usepackage{xcolor}

\usepackage{microtype}
\usepackage{graphicx}
\usepackage{booktabs} %
\usepackage{times}
\usepackage{rotating}

\usepackage{amsmath,amsfonts,bm}

\def\eqref#1{equation~\ref{#1}}

\def\1{\bm{1}}

\DeclareMathAlphabet{\mathsfit}{\encodingdefault}{\sfdefault}{m}{sl}
\SetMathAlphabet{\mathsfit}{bold}{\encodingdefault}{\sfdefault}{bx}{n}

\newcommand{\KL}{D_{\mathrm{KL}}}

\DeclareMathOperator*{\argmax}{arg\,max}

\usepackage[utf8]{inputenc} %
\usepackage[T1]{fontenc}    %
\RequirePackage{doi}
\usepackage{booktabs}       %
\usepackage{amsfonts}       %
\usepackage{amsmath}
\usepackage{nicefrac}       %
\usepackage{microtype}
\usepackage{graphicx}
\usepackage{subfig}
\usepackage{enumitem}
\usepackage{wrapfig}
\usepackage{mathtools}
\usepackage{pgffor}
\usepackage{grffile}
\usepackage{caption}
\usepackage{tikz}
\usepackage{multirow}
\usepackage{subfiles}
\usepackage{etoolbox}
\usepackage{floatrow}    
\usepackage{empheq}
\usepackage{todonotes}

\usetikzlibrary{patterns}
\usetikzlibrary{bayesnet}
\usetikzlibrary{arrows,decorations.markings}
\tikzstyle{disent_latent} = [circle,pattern=north east lines, pattern color=black!20,draw=black,inner sep=1pt,
minimum size=20pt, font=\fontsize{10}{10}\selectfont, node distance=1]

\DeclareMathOperator{\expect}{\mathbb{E}}

\DeclareMathOperator{\ELBO}{\mathcal{L}}

\DeclareMathOperator{\cn}{\mathcal{N}}

\DeclareMathOperator{\cat}{\mathrm{Cat}}

\newcommand{\vecto}[1]{\boldsymbol{\mathbf{#1}}}
\renewcommand{\v}{\vecto}

\begin{document}
\title{Semi-\textit{Un}supervised Learning: Clustering and
Classifying using Ultra-Sparse Labels}

\author{\IEEEauthorblockN{Matthew Willetts}
\IEEEauthorblockA{\textit{University of Oxford} \\
\textit{Alan Turing Institute}\\
mwilletts@turing.ac.uk}
\and
\IEEEauthorblockN{Stephen Roberts}
\IEEEauthorblockA{\textit{University of Oxford} \\
\textit{Alan Turing Institute}\\
sjrob@robots.ox.ac.uk}
\and
\IEEEauthorblockN{Chris Holmes}
\IEEEauthorblockA{\textit{University of Oxford} \\
\textit{Alan Turing Institute}\\
cholmes@stats.ox.ac.uk}
}

\maketitle

\IEEEpubidadjcol

\begin{abstract}
In semi-supervised learning for classification, it is assumed that every ground truth class of data is present in the small labelled dataset.
In many real-world sparsely-labelled datasets, it is possible that not all ground-truth classes are captured in the labelled dataset:
a biased data collection process could result in some classes of data to be found only in the unlabelled dataset.
We call this regime \textit{semi-unsupervised learning}, an extreme case of semi-supervised learning, where some classes have no labelled exemplars.
First, we outline the pitfalls associated with trying to apply deep generative model (DGM)-based semi-supervised learning algorithms to datasets of this type.  We then show how a combination of clustering and semi-supervised learning, using DGMs, can be brought to bear on this problem. We study several different datasets, showing how one can still learn effectively when half of the ground truth classes are entirely unlabelled and the other half are sparsely labelled.

\end{abstract}


\section{Introduction}
\label{intro}

What are the standard learning regimes when handling classes?
When learning to classify we can either perform supervised learning, if we have fully labelled data, or semi-supervised learning, if that labelling is sparse.
Or we can perform clustering via unsupervised learning if there is no labelled data at all.
Semi-supervised learning is useful as in many problem domains we have only a relatively small amount of labelled data compared to the amount of unlabelled data.

Gathering labels often requires expert annotation, which is expensive, whereas unsupervised data can often be obtained by automated methods.
Thus, it is common for only a subset of data gathered to be (expensively) labelled.
Within this sparsely-labelled dataset, however, it is possible that there may well be ground truth classes of data for which we have no labelled examples at all: some ground-truth classes of data are found only in the unlabelled dataset.
This would be most likely to be due to selection bias, where the labelled data is from a biased sample of the overall data distribution.
It is this variety of data that we consider here, focusing on image data.

First, we outline the pitfalls associated with trying to apply deep generative model (DGM)-based semi-supervised learning algorithms to datasets of this type.  We then show how a combination of clustering and semi-supervised learning, using DGMs, can be brought to bear on this problem. We study several different datasets, showing how one can still learn effectively when half of the ground truth classes are entirely unlabelled and the other half are sparsely labelled.

A hypothetical example of a dataset of this type would be a set of medical images, scans say.
As labelling is expensive, we obtain expert labelling -- the variety of pathology present -- for only some small proportion of all the scans we have.
It is plausible that we might not happen to capture all distinct types of pathology in this smaller labelled dataset: either because they are very rare or because the process for gathering the labels was biased in some way.
The latter could occur if we gathered all of our data to be labelled from only one hospital or ward that had a non-representative sample of the population.

Datashift could also cause a variety of biasing in the data collection.
If new varieties or behaviours emerge over time, and unlabelled data continues to be gathered cheaply, while the expensive process of data labelling is not continued, then newly-emergent classes of data will not be found in the labelled dataset, only the unlabelled.

In a dataset of this type, an unlabelled image could be from one of the varieties that is captured in the labelled dataset, or it could be of another, unseen, variety.
We would not be in the semi-supervised regime.
Nor do we want to treat the problem as unsupervised, discarding our limited yet still information-rich labelled data.

We call this \textit{semi-unsupervised learning}: we wish to jointly perform semi-supervised learning on sparsely-labelled classes and unsupervised learning on completely unlabelled classes.
This requires a model that can learn successfully in both unsupervised and semi-supervised regimes.

We study how learning on sparsely-labelled data of this type introduces challenges, and describe how those challenges can be answered.
First we discuss potential issues from using semi-supervised methods when the training data does not contain labelled examples from all ground truth classes -- ie when it is semi-\textit{un}supervised data.
We demonstrate this using deep generative models used for semi-supervised learning: one can end up attributing all data merely to the classes represented in the labelled dataset.
Then we show how some obvious approaches one might take to make use of semi-supervised learning algorithms do not work, even when one knows that the training data is, in fact, semi-unsupervised and tries to make allowances.
Finally, we show how certain varieties of deep generative model-derived clustering algorithms can be used to handle this learning regime, over a range of datasets.
In the case that half of the ground truth classes are masked out entirely, and the remaining half of classes are sparsely-labelled, we show we can still learn a classifier that performs well on the test set containing all ground-truth classes.

\section{Background}

\subsection{Semi-Supervised Learning}

If we had a fully labelled dataset we could easily train a discriminative model under maximum likelihood.
We have input data and (one-hot) labels drawn from a dataset $(\v{x},y)\sim(\v{X}_\ell,\v{Y}_\ell)=\mathcal{D}_\ell$. $\v{x}\in \mathbb{R}^{d_x}$ and $y\in\{0,1\}^K, \sum_{i=1}^K y_i =1$.
We could specify a parametric model for $p_\lambda(y|\v{x})$, say a deep neural network, and aim to find the optimal parameters
\begin{equation}
    \lambda^* = \argmax_\lambda \expect_{(\v{x},y)\sim\mathcal{D}}\log p_\lambda(y|\v{x})
    \label{eq:xent}
\end{equation}
via optimisation, say via stochastic gradient ascent on minibatches of data.
But in semi-supervised learning we have an additional dataset $\mathcal{D}_\mathrm{u}$, which contains unlabelled data $\v{x}$ with no corresponding label, and further $|\mathcal{D}_\mathrm{u}|\gg|\mathcal{D}_\ell|$.

Semi-supervised models for classification are designed to be able to learn from both datasets -- to learn from sparsely-labelled data.
This is of particular importance when $\mathcal{D}_\ell$ is too small to train a good enough classifier on its own.
The hope would be that by somehow extracting information from $\mathcal{D}_\mathrm{u}$ we can make a better classifier, say in terms of accuracy, when applied to new data, than if we had trained on $\mathcal{D}_\ell$ only \cite{SSbook}.

While there are numerous approaches one could take to try to extract information from the unlabelled data to train a better classifier, here we will focus on the use of probabilistic generative models.
As discussed in the introduction, in deep generative models the parameters of the distributions within the model are themselves parameterised by neural networks.
Due to the coherency of probabilistic modelling, these models can handle missing observations in a principled way.
Within this framework, we can do partial conditioning to obtain distributions of importance to us, here the (approximate) posterior distributions for unobserved labels.

\textbf{Generative Classification}
A simple generative model based approach would be to fit a parametric joint distribution over data $(\v{X}_\ell,\v{Y}_\ell) = \mathcal{D}_\ell$, and $\v{X}_\mathrm{u} = \mathcal{D}_\mathrm{u}$, then obtain a posterior for the unknown labels conditioned on all the data one has.
So we would first aim to find the maximum likelihood setting of the model parameters,
\begin{align}
    \theta^* &= \argmax_\theta p_\theta(\v{X}_\ell,\v{Y}_\ell, \v{X}_\mathrm{u}) \nonumber \\
    &= \argmax_\theta\sum_{\v{Y}_{u}^{'}} p_{\theta}(\v{X}_\mathrm{u}, \v{Y}_{u}^{'},  \v{X}_\ell, \v{Y}_\ell).
\end{align}
This requires summing over all $K^{|\mathcal{D}_\mathrm{u}|}$ possible arrangements of $\v{Y}_\mathrm{u}$.
Then find the posterior for $\v{Y}_\mathrm{u}$,
\begin{align}
p_{\theta^*}(\v{Y}_\mathrm{u} | \v{X}_\ell, \v{Y}_\ell, \v{X}_\mathrm{u}) = \frac{p_{\theta^*}(\v{X}_\mathrm{u}, \v{Y}_\mathrm{u},  \v{X}_\ell, \v{Y}_\ell)}{\sum_{\v{Y}_{u}^{'}} p_{\theta^*}(\v{X}_\mathrm{u}, \v{Y}_{u}^{'},  \v{X}_\ell, \v{Y}_\ell)},
\label{eq:genclass}
\end{align}
where again we have to perform  a sum with a geometrically-scaling number of terms.
Reminiscent of \cite{Blei2017a}, the integrals required to compute this posterior are often intractable for large datasets and large label spaces, so we must make approximations so that we can learn efficiently.
And it is desirable that our learning algorithm results in a classifier as a \textit{inference artifact}, that can be called on new data just as a normal neural network classifier can be.

\subsection{Semi-Supervised Variational Autoencoders}
We attack this problem using Variational auto-encoders (VAEs) \cite{Kingma2013, Rezende2014} as they offer a modelling paradigm that can give us both of these properties.
The semi-supervised VAE (SSVAE) proposed in \cite{Kingma2014a} is a simple extension for semi-supervised learning.
It has a continuous latent variable $\v{z}\in \mathbb{R}^{d_z}$ and a partially-observed class variable $y$.
For $\mathcal{D}_\mathrm{u}$ we only have input data $\v{x}$, so for each $\v{x}$ there is a corresponding latent variable $y$.
For $\mathcal{D}_\ell$ we have observed $y$, so $y$ is not a latent variable.
In \cite{Kingma2014a} the joint distribution is:
\begin{align}
p_{\theta}(\v{x},y,\v{z})&= p_{\theta}(\v{x}|y,\v{z})p(y)p(\v{z}) \label{eq:m21}
\end{align}
where $p(y) = \cat(y|\pi)$ and $p(\v{z})=\cn(\v{z}|0,\mathbb{I})$.
$p_\theta(\v{x}|y,\v{z})$ is an appropriate distribution given the form of the data.
For continuous data it is commonly a Gaussian with fixed diagonal covariance: $p_{\theta}(\v{x}|y, \v{z}) = \mathcal{N}(\v{x}|\mu_{\theta}(\v{z},y),\Sigma_{\theta})$, with
$\mu_{\theta}(\v{z},y)$ parameterised by a neural network.
For binary data a set of Bernoulli distributions is common.

As exact inference is intractable, we perform stochastic amortised variational inference.
We aim to optimise a lower bound on the evidence for all our data, the two datasets $\mathcal{D}_\mathrm{u}, \mathcal{D}_\ell$.
In each case, we need a variational posterior for the unobserved variables.
$\v{z}$ is always latent, and $y$ is sometimes-latent, sometimes-observed.
So for semi-supervised data the evidence lower bound consists of two terms.
First, for unlabelled data ($y$ is a latent variable to be inferred),
\begin{equation}
\ELBO_\mathrm{u}(\v{x})= \expect_{\v{z},y \sim q_\phi(\v{z}, y|\v{x})}\log \frac{p_{\theta}(\v{x},y,\v{z})}{q_\phi(\v{z}, y|\v{x})}.
\label{eq:unlabelbo}
\end{equation}
Secondly, for labelled data ($y$ is observed),
\begin{equation}
\ELBO_\ell(\v{x}, y)= \expect_{\v{z} \sim q_\phi(\v{z}| \v{x},y)}\log \frac{p_{\theta}(\v{x},y,\v{z})}{q_\phi(\v{z}| \v{x},y)}.
\label{eq:labelbo}
\end{equation}
Choosing a particular form for the posterior:
\begin{align}
q_{\phi}(\v{z},y|\v{x})&=q_{\phi}(\v{z}|\v{x},y)q_{\phi}(y|\v{x}) \label{eq:m22}\\
q_{\phi}(y|\v{x})&=\cat(y|\pi_{\phi}(\v{x})) \label{eq:m23}\\
q_{\phi}(\v{z}|\v{x},y)&=\mathcal{N}(\v{z}|\mu_{\phi}(\v{x},y),\Sigma_{\phi}(\v{x},y)) \label{eq:m24}
\end{align}
where $\mu_{\phi}(\v{x},y),\Sigma_{\phi}(\v{x},y), \pi_{\phi}(\v{x})$ are neural networks.

$q_{\phi}(y|\v{x})$ is an inference artifact, a classifier.
But note that it only appears in $\ELBO_\mathrm{u}(\v{x})$, so it would only be trained on unlabelled data, clearly an undesirable property for a classifier.
To remedy this, add to $\ELBO_\ell(\v{x})$ the cross entropy classifier loss, the same as inside Eq (\ref{eq:xent}), weighted by a factor $\alpha$ \cite{Kingma2014a}.
The overall objective with unlabelled data $\mathcal{D}_\mathrm{u}$ and labelled data $\mathcal{D}_\ell$ is the sum of the evidence lower bounds for all data with this classification loss:
\begin{align}
\ELBO(\mathcal{D}_\mathrm{u}, \mathcal{D}_\ell) =  \smashoperator{\expect_{(\v{x}_\ell,y_\ell) \sim \mathcal{D}_\ell}} &[\ELBO_\ell(\v{x}_\ell,y_\ell) - \alpha  (\log q_{\phi}(y_\ell |\v{x}_\ell))] + \smashoperator{\expect_{\v{x}_\mathrm{u} \sim \mathcal{D}_\mathrm{u}}} \ELBO_\mathrm{u}(\v{x}_\mathrm{u}) \label{eq:total_loss}
\end{align}
Through joint optimisation over $\{\theta, \phi\}$ using stochastic gradient ascent we aim to find point-estimates of those parameters that maximises the evidence lower bound over both our datasets $\mathcal{D}_\ell$ and $\mathcal{D}_\mathrm{u}$.

For the expectations over $\v{z}$ and $y$ in the objective, we take Monte Carlo (MC) samples from the variational posteriors.
To take derivatives through these samples wrt $\theta, \phi$ use \textit{reparameterisation tricks}.
For $\v{z}$ this means rewriting the sample as a deterministic function given a sample from $\mathcal{N}(0,\mathbb{I})$:
\begin{equation}
\v{z}\sim\mathcal{N}(\v{z}|\v{\mu},\v{\Sigma}) \Longleftrightarrow \v{\epsilon}\sim\mathcal{N}(\v{\epsilon}|0,\mathbb{I}), \v{z}=\v{\mu} + \v{\Sigma}^\frac{1}{2} \cdot \v{\epsilon}
\end{equation}
and for $y$ using the Gumbel-Softmax Trick/CONCRETE sampling \cite{Maddison2016, Jang2016}.

\section{Semi-Unsupervised Data with Semi-Supervised Models}
\label{sec:sus_w_ss}
Can we use standard semi-supervised approaches with semi-\textit{un}supervised data?
In this section we demonstrate how SSVAEs perform.
First, we cover the \textit{by accident} case where we think we are performing semi-supervised learning, but our unlabelled data does contain extra classes.
Then we cover the \textit{on purpose} case, where we expect that there are extra classes of data and make reasonable allowances for them in our semi-supervised modelling.

\subsection{By Accident}
What happens if we train an SSVAE described above, on data that we believe conforms to our requirements for semi-supervised learning, but in fact contains additional ground-truth classes of data in the unlabelled dataset?

\textbf{Experiments}
To mimic semi-unsupervised data using standard datasets we simply mask out values of the training data.
We train an SSVAE on MNIST \cite{lecun2010} and Fashion-MNIST \cite{Xiao2017}.
For these experiments, we have to pick the overall dimensionalities of $d_z$.
We use $d_z=5$ for MNIST and $d_z=10$ for Fashion-MNIST.
We for this and all subsequent experiments we use small MLPs to represent the parameters for the distributions in the generative and recognition models.
We train our models using ADAM \cite{Kingma2015}.
For full implementation details, see the appendix.

The labelled training set $\mathcal{D}_\ell$ contains examples only of classes $\{0,..,4\}$, 20\% of the total examples in the standard training set, so $\approx1000$ labelled examples for each class in (Fashion-)MNIST.
The unlabelled dataset $\mathcal{D}_\mathrm{u}$ is thus all other training data, with labels dropped: all training data points for classes $\{5,..9\}$ and the left-over 80\% of training data points for classes $\{0,..,4\}$.

We compare this to being trained in an equivalent semi-supervised manner, where the unlabelled data just contains classes $\{0,..,4\}$.
We find that for MNIST and Fashion-MNIST there is no significant change in test-set classification performance on classes $\{0,..,4\}$ due to the additional presence of unexpected classes in the unlabelled data, see Table \ref{tab:accidental_results}.

\begin{figure}
    \centering
    \makebox[0.5\textwidth]{
    \subfloat[MNIST - SS]{
    \includegraphics[width=3.5cm]{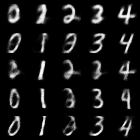}
    }
    \subfloat[MNIST - SUS accident]{
    \includegraphics[width=3.5cm]{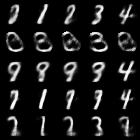}
    }
    }
    \makebox[0.5\textwidth]{
    \subfloat[F-MNIST - SS]{
    \includegraphics[width=3.5cm]{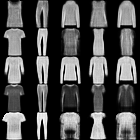}
    }
    \subfloat[FMNIST - SUS accident]{
    \includegraphics[width=3.5cm]{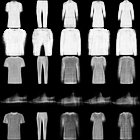}
    }
    }
    \caption{We show the effect of having unexpected additional classes in the training set for an SSVAE when trained on MNIST and Fashion-MNIST. We are plotting the mean of $p_\theta(\v{x}|y,\v{z})$ for a range of samples of $\v{z}$ and for all values of $y$. In each plot, each row is generated using a shared draw $\v{z}^*\sim p(\v{z})$, with the columns indexing over $y$. The models were trained with a small labelled dataset of classes $\{0,...,4\}$. The semi-supervised (SS) model's unlabelled training set contained only those same classes. The semi-unsupervised by accident (SUS accident) model's unlabelled training set contained all classes $\{0,...,9\}$. The \textit{semi-supervised} plots a) and c) shows successful controlled generation, as each column -- each value of y -- corresponds to a distinct class, with $\v{z}$ encoding style. The \textit{semi-unsupervised by accident} plots b) and d) show that for some settings of $\v{z}$ we get generation of datapoints that look like the unlabelled-only classes, in some cases regardless of the value of $y$: in b) the $2^{\mathrm{nd}}$ and $3^{\mathrm{rd}}$ rows, in d) the $4^{\mathrm{th}}$ row.}
    \label{fig:ss_sus_gen}
\vspace{-1mm}
\end{figure}

\begin{table}[t]
\caption{Accidental Semi-Unsupervised Learning. We want to see how performance changes when a semi-supervised model is trained with additional unlabelled classes in the unlabelled dataset. We show here the test set accuracy (Acc. $\%\pm \mathrm{SD}$) of SSVAE trained on MNIST and Fashion-MNIST (F-MNIST) over four runs when trained in two ways. SS (semi-supervised): trained with vanilla semi-supervised data but only for classes $\{0,..,4\}$ with 20\% of datapoints labelled. Accidental SUS (semi-unsupervised): the unlabelled training set also includes unlabelled examples of classes $\{5,...,9\}$.
The test set is classes $\{0,..,4\}$ only in both cases.}
\begin{center}
\begin{small}
\begin{sc}
\begin{tabular}{ccc}
\toprule
Dataset&    SS                     & Accidental SUS\\ \midrule
MNIST  &    $97.0 \pm 0.4$         & $97.4 \pm 0.3$ \\
\midrule
F-MNIST&    $90.1 \pm 0.3$     & $89.8 \pm 0.2$ \\
\bottomrule

\end{tabular}
\end{sc}
\end{small}
\end{center}
\label{tab:accidental_results}
\end{table}
More apparent is the the effect on generated data, as semi-unsupervised data distorts the properties of the forward model.
Under maximum-likelihood training, models with sufficient capacity will learn a way to have representations that correspond to the unlabelled-only classes.
Put another way, with sufficiently powerful neural architectures these models will learn to reconstruct all their training data with high fidelity, meaning that there will be some setting of their latent representations that correspond closely to any given training datapoint.
Thus, when sampling the values of the model's latents during generation, those settings that correspond to unlabelled-only classes can be picked.
We see this in Figure~\ref{fig:ss_sus_gen}.
We sample from the model for a set of samples $\v{z}^*\sim p(\v{z})$, plotting the mean of $p(\v{x}|\v{z}^*,y)$ for each value of $y$.
Each row corresponds to the same $\v{z}^*$ sample, each column corresponding to a value of $y$.
For the models trained on semi-supervised data, generation is well controlled by these variables, with $y$ controlling identity and $\v{z}$ controlling style.
For the models trained on semi-unsupervised data, we sometimes generate data that looks like one of the only-unlabelled classes $\{5,...,9\}$.
By the design of the model, this cannot be done for a particular setting of $y$, so instead this is done through the information in $\v{z}$.
Thus we see that for some samples of $\v{z}$, regardless of the value of $y$, the decoder output looks like an image from one of the unlabelled-only classes.

\begin{figure}
    \centering
    \makebox[0.5\textwidth]{
    \hspace{-0.3cm}
    \subfloat[MNIST]{
    \includegraphics[width=7cm]{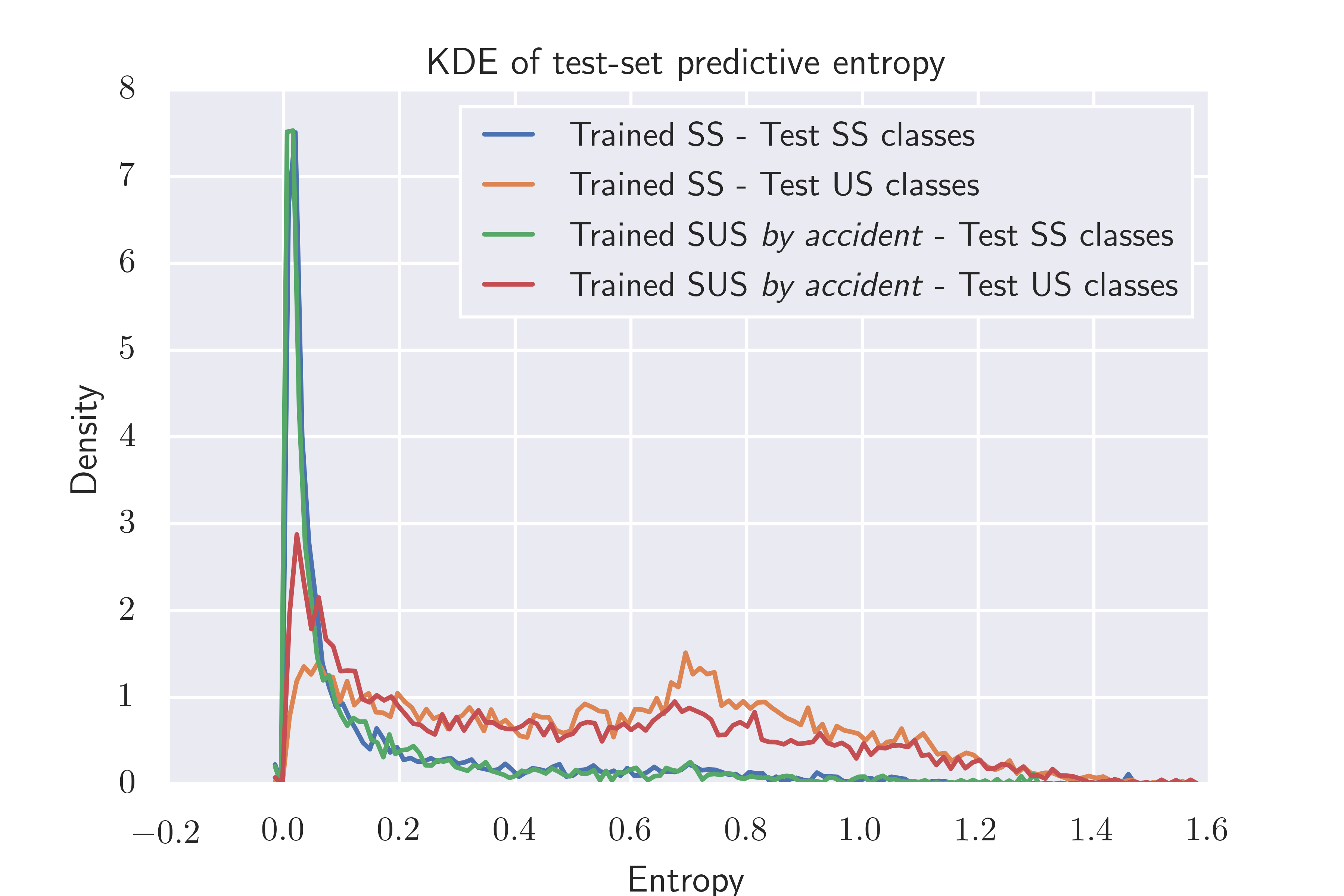}
    }}
    \makebox[0.5\textwidth]{
    \subfloat[F-MNIST]{
    \includegraphics[width=7cm]{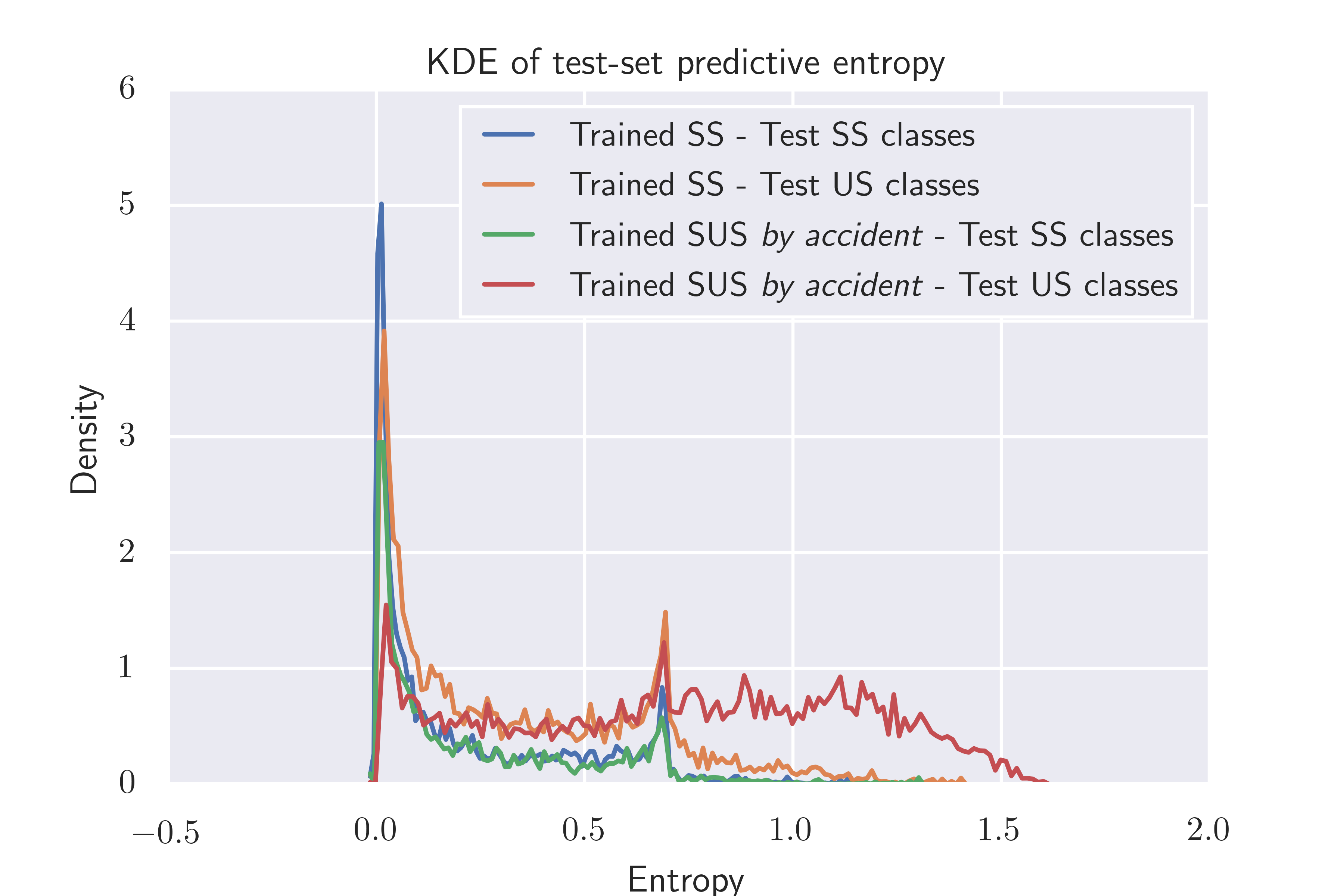}
    }
    }
    \caption{KDE plots of entropy of test set $q_\phi(y|\v{x})$ for MNIST and F-MNIST with the same training setup as Figure~\ref{fig:ss_sus_gen}. The test set is either semi-supervised classes (SS) $\{0,...,4\}$ only or unsupervised classes (US) $\{5,...,9\}$ only. We see that for F-MNIST when trained SUS \textit{by accident} there is higher entropy for the US classes, but for MNIST there is not a clear effect.}
    \label{fig:qy_hist}
\vspace{-1mm}
\end{figure}
While it is perhaps desirable that a semi-supervised model's classification performance is robust to having been trained with semi-unsupervised data, it means that we cannot use that performance metric to detect if we have erred in our assumption that we are in the semi-supervised regime.

The main problems with training a semi-supervised model on semi-unsupervised data \textit{by accident} are first that a practitioner may struggle to know that there are extra classes hidden in their unlabelled dataset, and second, the model will necessarily attribute all incoming data point to only the classes in the limited labelled dataset.

\subsection{On Purpose}
\label{sec:on_purpose}
If we think there are additional ground-truth classes present in the unlabelled dataset $\mathcal{D}_\mathrm{u}$, a natural approach would be to have extra dimensions in $y$.
For example, if we have $n_\ell$ classes each with some labelled examples and we think there is some number $n_\mathrm{aug}$ of additional classes present in the unlabelled data, we could reasonably choose $y\in\{0,...,(n_\ell+n_\mathrm{aug})\}$.
We might then hope that our semi-supervised learning algorithm would make use of these `empty' dimensions in $y$, using them to encode the additional classes we hope it will discover.

Further, we could choose $n_\mathrm{aug}$ to be more than the number of extra ground-truth classes we think there are, giving more components in $y$ over to them: a common practice in clustering \cite{Yang2016, Makhzani2016, Kilinc2018, Dilokthanakula}.
It gives us a degree of insurance against sub-optimal agglomeration of different kinds of data into discrete units within the model.
For example, for MNIST we might think that we should give the model the potential to use multiple values of $y$ for the same ground-truth class to capture the distinct varieties that exist in how people write certain digits.
Sticking with the above, masked, dataset, from having access to labels for classes $\{0,..,4\}$ we might have noticed that people write $4$s in two different manners.
So any extra classes of digit we might discover in the unlabelled data may also vary in their manner of writing.

\textbf{Prior in $y$} We need to specify a prior $p(y)$.
We divide the probability mass between the known-classes and the expected-classes in proportion to our prior expectation of their proportions in the data.
Then within the known, labelled classes we choose a prior proportional to their frequency; for the expected additional classes, we divide mass uniformly.

For the datasets we study, the ground truth classes are of approximately equal frequency.
So we divide the mass of our prior in half between the classes we have some labelled examples for and the expected classes.
Thus
\[
p_\theta(y)=
\begin{cases}
\frac{1}{2 n_\ell} & y \in \{0, ... ,n_\ell\} \\
\frac{1}{2 n_{\mathrm{aug}}} & y \in \{n_\ell + 1, ... ,K\}
\label{eq:prior}
\end{cases}
\]
where $K=n_{\mathrm{aug}}+n_\ell$.
We use this prior for all experiments where we are trying to take account of the presence of additional unlabelled classes.

\textbf{Evaluating performance on unlabelled classes}
In evaluating a semi-unsupervised model, we must choose a method for evaluation that naturally applies both to the semi-supervised and the unsupervised classes.
In pure semi-supervised learning, we can simply measure accuracy on the test set.
A close analogy to accuracy, that is used to evaluate DGM-based clustering algorithms is cluster accuracy (ACC) \cite{Jiang2017, Yang2016, Kilinc2018}.
This is a test time version of \textit{cluster and label}.
\begin{equation}
    \mathrm{ACC} = \max_{P \in \mathcal{P}}\frac{\sum_{i=1}^{|\mathcal{D}|}\mathbb{I}[t_i=P y_i]}{|\mathcal{D}|}
\end{equation}
where $P$ is a $T \times K$ rectangular permutation matrix that attributes each $y$ to a ground truth class $t$.

We can think of this as solving an assignment problem.
We need to link each of the learnt, unsupervised classes with a particular (never seen labelled during training) ground-truth class.
So we attribute each learnt class to the ground truth class that is the most common ground truth class within it at test time.
The assignment we get from this method will also enable us to plot confusion matrices in the semi-unsupervised case.

\textbf{Experiments}
We use Fashion-MNIST, MNIST, and HAR \cite{Stisen2015}; masking the first half of classes partially and the second half fully in the same way as in the sub-section above.
We augment $K = n_\ell + n_{\mathrm{aug}} = n_\mathrm{gt}/2 + n_{\mathrm{aug}}$.
We use $n_\mathrm{aug}=40$ in these experiments.
We see from the confusion matrices in Figure~\ref{fig:m2_conf} and the overall results in Table \ref{tab:results} that no informative classifier is learnt over the unlabelled classes by an SSVAE.

\subsection{Inductive Bias Matching}
\label{subsec: pc}

\begin{figure}[h!]
    \centering
    \makebox[0.5\textwidth]{
    \subfloat[Samples from clustering SSVAE]{
    \includegraphics[width=7cm]{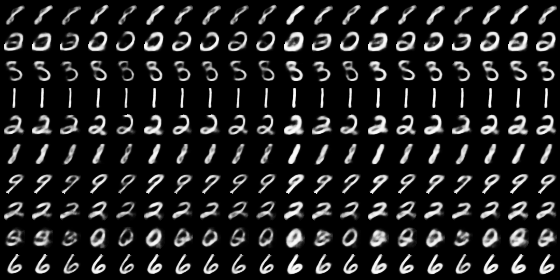}
    }
    }
    \makebox[0.5\textwidth]{
    \subfloat[SSVAE's most confidently assigned data-points]{
    \includegraphics[width=7cm]{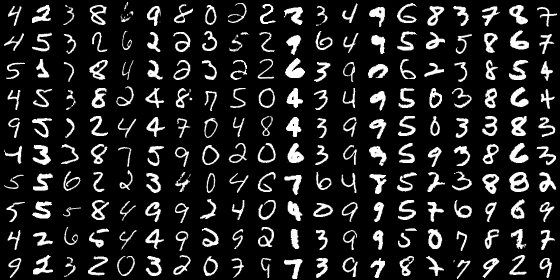}
    }
    }
    \caption{We trained an SSVAE unsupervised with a 20 dimensional discrete latent $y$. In a), as in Figure 1, we are showing the mean of $p_\theta(\v{x}|y,\v{z})$ conditioned on all values of $y$ for various samples of $\v{z}$ drawn from the prior. Columns index over values of $y$, each row being a particular $\v{z}$ samples.
    In b) we show the 10 most confidently-assigned data points in the test set for this model. From both a) and b) we see that different $y$ classes, different columns, correspond to different stroke thicknesses, and from a) we see that digit identity is represented in $\v{z}$.}
    \label{fig:inductive_bias}
\vspace{-1mm}
\end{figure}

In an SSVAE, we are specifying that the latent space is composed of a continuous part and a discrete part.
By training on semi-supervised data, we direct the discrete part to correspond to the limited class labels that we have and that the continuous part should encode residual `style' information.

By construction, generative models used for semi-supervised learning can be run using only unlabelled data: nothing is stopping one from using such a model for clustering.
When trained on unlabelled information we can see that an SSVAE does not learn a clustering that matches the ground-truth classes we are interested in.
On MNIST, the assignment of roles between the two latent variables is reversed: $\v{z}$ encodes identity, with $y$ encoding stroke thickness.
We see this in Figure~\ref{fig:inductive_bias}, where we show samples from an SSVAE trained unsupervised on MNIST with 20 cluster components.
Each row corresponds to a sample from $p(\v{z})$ and each column to a setting of $y$.
This carries over to the unlabelled sub-problem in semi-unsupervised learning: we have poor performance over the unlabelled classes as we can see in Figure~\ref{fig:m2_conf} and Table \ref{tab:results}.
One might expect that the presence of some labelled data would induce the appropriate division of roles between $\v{z}$ and $y$, with all (labelled and unlabelled) classes indexed in $y$, but this is not the case.
Instead, we must turn to approaches with the appropriate clustering performance to solve this task.

\section{Semi-Unsupervised Data with Clustering Models}

If we attempt to learn using semi-unsupervised data with semi-supervised models with the wrong inductive biases we run into problems.
Even when expanding the dimensionality of the discrete label-space, we find that for some deep probabilistic generative semi-supervised learning algorithms the model will not use these additional components to solve the task we care about.
Classes of data found only in the unlabelled dataset are not separated.
This is because these models can not perform clustering in the terms we care about, including when that clustering is a sub-problem.

\begin{table*}
\caption{Test set accuracy/cluster purity of SSVAE and GM-DGM trained on MNIST, Fashion-MNIST (F-MNIST) and HAR, over four runs each for unsupervised learning (US), semi-supervised learning (SS) and semi-unsupervised learning (SUS).}
\begin{center}
\begin{small}
\begin{sc}
\begin{tabular}{cclll}
\toprule
Model&Dataset& US Acc. $\%\pm \mathrm{SD}$& SS Acc. $\%\pm \mathrm{SD}$& SUS Acc. $\%\pm \mathrm{SD}$ \\ \midrule
SSVAE      &MNIST  & $26.1 \pm 2.8$  & $97.6 \pm 0.1$  & $54.3 \pm 0.9$ \\
GM-DGM  &MNIST  & $\v{90.0 \pm 1.5}$  & $\v{97.7 \pm 0.1}$  & $\v{92.5 \pm 0.3}$ \\
\midrule
SSVAE      &F-MNIST& $18.2 \pm 0.4$  & $86.8 \pm 0.2$  & $38.7 \pm 1.7$ \\
GM-DGM  &F-MNIST& $\v{75.8 \pm 0.3}$  & $\v{86.9 \pm 0.1}$  & $\v{78.22 \pm 1.1}$ \\
\midrule
SSVAE       &HAR& $29.0 \pm 1.1$  & $\v{97.7 \pm 0.1}$  & $49.7 \pm 7.5$ \\
GM-DGM  &HAR& $\v{81.7 \pm 1.8}$  & $96.6 \pm 0.3$  & $\v{87.1 \pm 2.3}$ \\
\bottomrule

\end{tabular}
\end{sc}
\end{small}
\end{center}
\label{tab:results}
\end{table*}

Thus a reasonable way to wish to learn with semi-unsupervised datasets is to have a model that, in the absence of any label information, learns to cluster our data in a way that corresponds to the limited label information we do have.
When we train our model with semi-unsupervised data, we expect those additional classes to have a similar nature to the partially-labelled classes.
If our model clusters appropriately, we can expect it to assign any new classes of data (present only in the unlabelled dataset) to distinct components in $y$.

This paper has used SSVAEs as an anchoring example, and it demonstrates a general point around using semi-supervised models for, in effect, clustering.
We cannot expect that generative models used for semi-supervised learning will necessarily have the right inductive biases to be run as appropriate clustering algorithms for the particular kind of label information we are interested in.
For that reason, in order to capture semi-unsupervised learning, we have to have an appropriate clustering algorithm.
We call this the \textit{inductive bias requirement}: we want to have a clustering algorithm that has the right inductive bias to produce clustering at the right level for our task -- digit identity rather than stroke thickness, for MNIST.
This property is not given by an SSVAE, for any of the datasets we have studied.

Further, if we are working with VAE-derived models, we are aided if our model has an amortised posterior for the clustering latent variable.
For then we can train powerful neural network classifiers inside our models.
We call this the \textit{classifier requirement}.
This property is given by an SSVAE.

Our experiments above tell us that we were unlucky in trying to capture semi-unsupervised learning with an SSVAE, as our task of encoding digit identity or garment-type into $y$ is not the natural mode of behaviour for this model. 
Conversely, if we had been interested in semi-unsupervised learning of discrete degrees of stroke width, say, then an SSVAE would be an appropriate model.
Moving forward with the standard labels given with the data sets under consideration, we show how a clustering algorithm that fulfills both the inductive-bias requirement and the classifier requirement can be used to perform semi-unsupervised learning.

\subsection{Gaussian Mixture Deep Generative Models}
\begin{figure}
\centering
\noindent\makebox[0.2 \textwidth]{%
\subfloat[][Gen.\\ $y$ latent]{\scalebox{0.6}{
\tikz{ %
      \node[obs] (x) {$\v{x}$} ; %
    \node[latent, above=of x] (z) {$\v{z}$} ; %
    \node[latent, above=of z] (y) {y} ; %
    \node[det, left=of z, , yshift=0cm] (theta) {$\theta$} ; %
    \plate[inner sep=0.4cm, xshift=0cm, yshift=0.12cm] {plate1} {(z) (x) (y)} {\scalebox{1}{$N_{\mathrm{u}}$}}; %
    \edge {theta} {z, x} ; %
    \edge {z} {x} ; %
        \edge {y} {z} ; %
  }}
  \label{fig:gm_dgm_generative1}}
  \hspace*{3mm}
  \subfloat[][Post.\\ $y$ latent.]{\scalebox{0.6}{{
\tikz{ %
      \node[obs] (x) {$\v{x}$} ; %
    \node[latent, above=of x] (z) {$\v{z}$} ; %
    \node[latent, above=of z] (y) {y} ; %
    \node[det, right=of z, , yshift=0cm] (phi) {$\phi$} ; %
    \plate[inner sep=0.4cm, xshift=0cm, yshift=0.12cm] {plate1} {(z) (x) (y)} {\scalebox{1}{$N_{\mathrm{u}}$}}; %
    \edge {phi} {z, y} ; %
    \edge {x} {z} ; %
    \edge {z} {y} ; %
      \draw [->] (x) to [out=115,in=-115] (y);
  }}
 \label{fig:gm_dgm_inf1}}}}
\hspace{22mm}
\noindent\makebox[0.2 \textwidth]{%
\subfloat[][Gen.\\ $y$ latent]{\scalebox{0.6}{
\tikz{ %
      \node[obs] (x) {$\v{x}$} ; %
    \node[latent, above=of x] (z) {$\v{z}$} ; %
    \node[obs, above=of z] (y) {y} ; %
    \node[det, left=of z, , yshift=0cm] (theta) {$\theta$} ; %
    \plate[inner sep=0.4cm, xshift=0cm, yshift=0.12cm] {plate1} {(z) (x) (y)} {\scalebox{1}{$N_{\ell}$}}; %
    \edge {theta} {z, x} ; %
    \edge {z} {x} ; %
        \edge {y} {z} ; %
  }}
  \label{fig:gm_dgm_generative2}}
  \hspace*{3mm}
  \subfloat[][Post.\\ $y$ obs.]{\scalebox{0.6}{\raisebox{0ex}{
\tikz{ %
      \node[obs] (x) {$\v{x}$} ; %
    \node[latent, above=of x] (z) {$\v{z}$} ; %
    \node[obs, above=of z] (y) {y} ; %
    \node[det, right=of z, , yshift=0cm] (phi) {$\phi$} ; %
    \plate[inner sep=0.4cm, xshift=0cm, yshift=0.12cm] {plate1} {(z) (x) (y)} {\scalebox{1}{$N_{\ell}$}}; %
    \edge {phi} {z} ; %
    \edge {x} {z} ; %
  }}
 \label{fig:gm_dgm_inf2}}}
 }
 \caption{Generative and Approximate Posterior models for GM-DGM, where $N_{\mathrm{u}}$ is the number of unlabelled points and $N_{\ell}$ the number of labelled points.}
 \label{fig:gm_dgm}
    \vspace{-1mm}
    \end{figure}
\label{sec:adding_mixtures}

Gaussian Mixture Deep Generative Models (GM-DGMs) have been popular clustering algorithms.
In various guises, they have been shown to produce clusters that correspond well to the ground-truth labels provided on various machine learning datasets \cite{Dilokthanakula, Jiang2017, Nalisnick2016, Shu2016}.
This means it fulfills the inductive bias requirement.

A simple version of a GM-DGM is to have a Gaussian mixture in $\v{z}$, each component the result of conditioning on $y$:
\begin{align}
p_{\theta}(\v{x},y,\v{z})&=p_{\theta}(\v{x}|\v{z})p_{\theta}(\v{z}|y)p(y) \label{eq:gm_dgm_gen}\\
p(y)&= \textrm{Cat}(y|\pi) \label{eq:gm_dgm_gen1} \\
p_{\theta}(\v{z}|y)&= \mathcal{N}(\v{z}|\v{\mu}_{\theta}(y),\v{\Sigma}_{\theta}(y)) \label{eq:gm_dgm_gen2}
\end{align}
And use the same form of variational posterior as used for SSVAEs, Eqs~(\ref{eq:m22} - \ref{eq:total_loss}).
By keeping that posterior, we keep a parameterised $q_\phi(y|\v{x})$, fulfilling the classifier requirement.
The training objective remains Eq~(\ref{eq:total_loss}), using the new generative model.
See Figure~\ref{fig:gm_dgm} for a graphical representation of this model.

\section{Experiments}
\label{sec:exp}

We keep the experimental setup as in Section \ref{sec:on_purpose}.
And as in earlier experiments, we use small MLPs to parameterise the parameters of the distributions in the generative and recognition models, trained using ADAM \cite{Kingma2015}.
Again we use $d_z=5$ for MNIST, $d_z=10$ for Fashion-MNIST, and $d_z=15$ for HAR.
We describe our model implementation and data processing in detail in Appendix C.

As before, in semi-supervised runs, we keep labels for $20\%$ of the data.
In the semi-unsupervised learning experiments, we keep 20$\%$ of label data for the first half of classes: $\{0,...,4\}$ for (Fashion-)MNIST, $\{0,1,2\}$ for HAR.
We then mask out all label information for the $n_{\mathrm{gt}}/2$ remaining classes.

\begin{figure}[h!]
    \vspace{1cm}
    \centering
    \subfloat[SSVAE Test Set Confusion Matrix on MNIST]{
    \includegraphics[width=8.1cm]{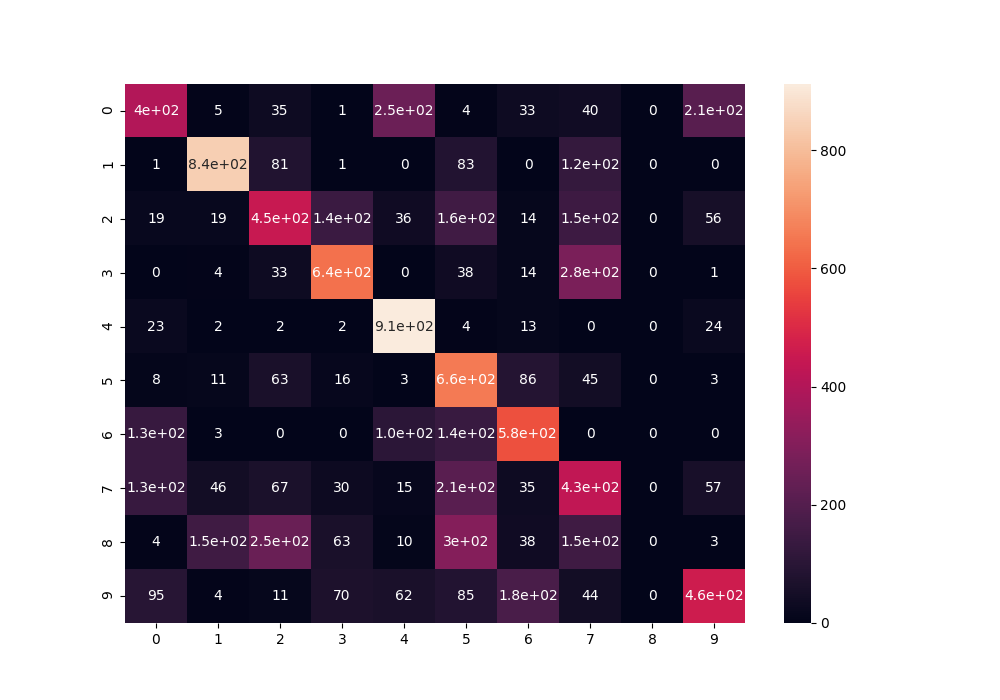}
    \label{fig:m2_mnist_conf_sus}
    }
    \vspace{1mm}
    \subfloat[SSVAE Test Set Confusion Matrix on F-MNIST]{
    \includegraphics[width=8.1cm]{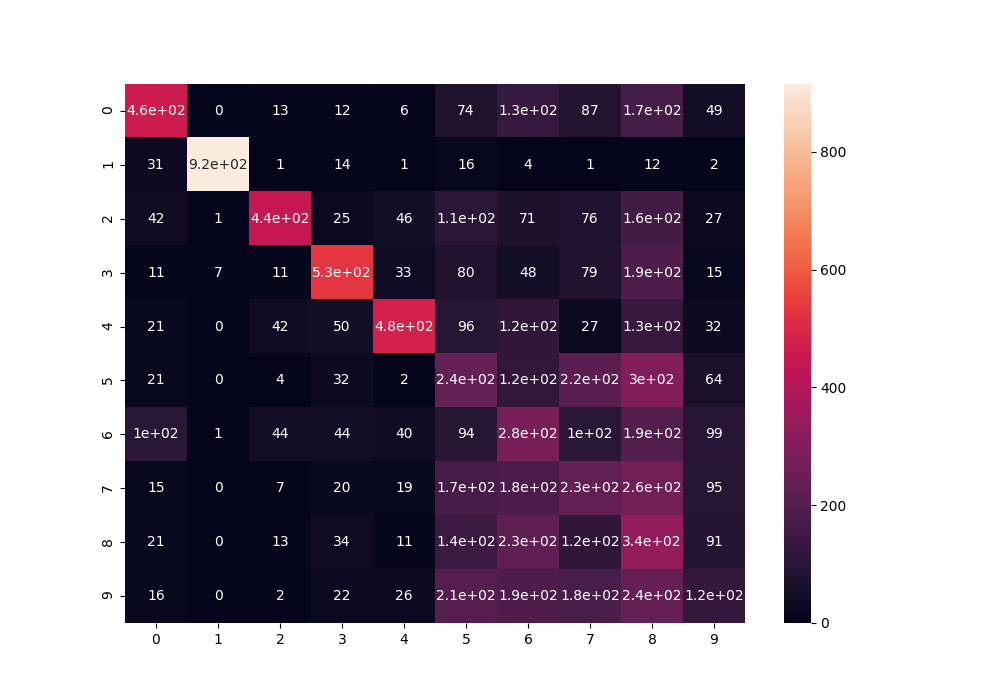}
    \label{fig:m2_fashion_conf_sus}
    }
    \vspace{1mm}
    \subfloat[SSVAE Test Set Confusion Matrix on HAR]{
    \includegraphics[width=8.1cm]{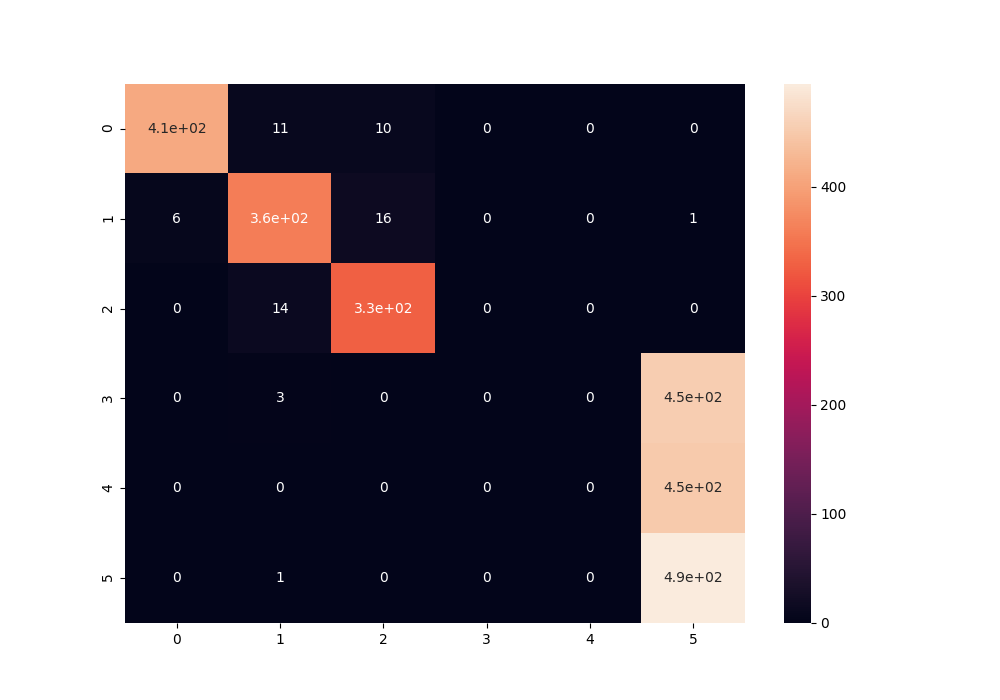}
    \label{fig:m2_har_conf_sus}
    }
\caption{Test set confusion matrices for SSVAEs trained semi-unsupervised on MNIST, Fasion-MNIST, and HAR}
\vspace{1cm}
\label{fig:m2_conf}
\end{figure}

\begin{figure}[h!]
\vspace{1cm}

    \centering
    \subfloat[GM-DGM Test Set Confusion Matrix on MNIST]{
    \includegraphics[width=8.1cm]{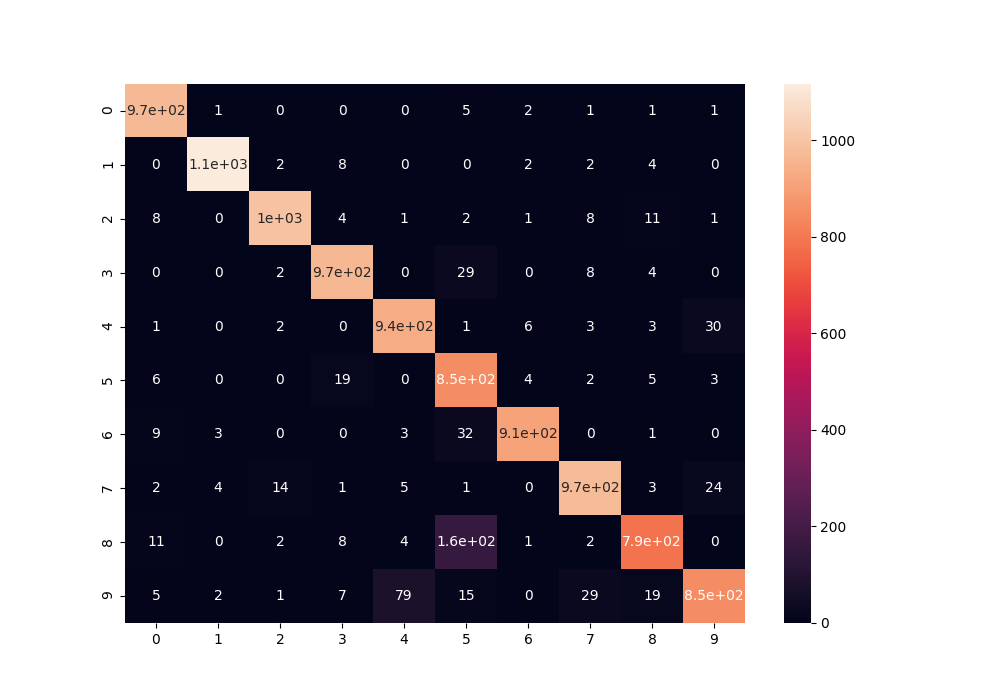}
    \label{fig:gm_dgm_mnist_conf_sus}
    }
    \vspace{1mm}
    \subfloat[GM-DGM Test Set Confusion Matrix on F-MNIST]{
    \includegraphics[width=8.1cm]{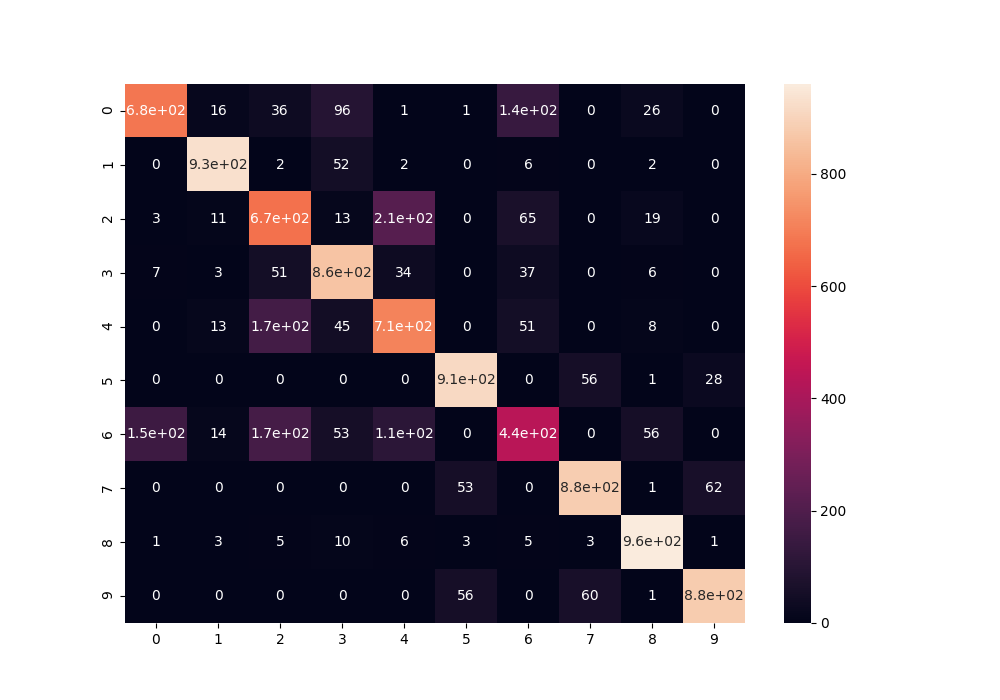}
    \label{fig:gm_dgm_fashion_conf_sus}
    }
    \vspace{1mm}
    \subfloat[GM-DGM Test Set Confusion Matrix on HAR]{
    \includegraphics[width=8.1cm]{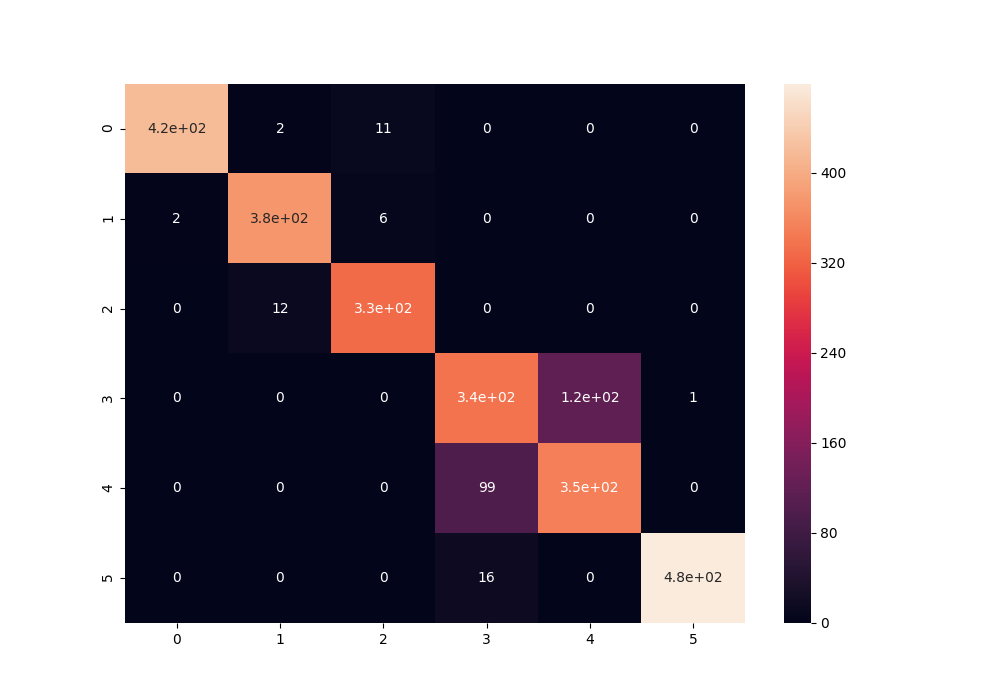}
    \label{fig:gm_dgm_har_conf_sus}
    }
    \caption{Test set confusion matrices for GM-DGMs trained semi-unsupervised on MNIST Fashion-MNIST and HAR.}
    \vspace{1cm}
\label{fig:gm_conf}
\end{figure}
Figure~\ref{fig:gm_conf} shows the confusion matrices for GM-DGMs trained semi-unsupervised on our studied datasets.
While the SSVAE was unable to cluster the subspace of unsupervised classes in $y$, as we saw in Figure~\ref{fig:m2_conf}, our GM-DGM model can learn a predictive classifier/clusterer for both the semi-supervised and unsupervised classes.
We show additional confusion matrices for SSVAEs and GM-DGMs trained unsupervised and semi-supervised on these datasets in the Appendix.

We show the test-set accuracy after unsupervised, semi-supervised, and semi-unsupervised learning for SSVAEs and GM-DGMs in Table \ref{tab:results}.
While SSVAEs demonstrably fail to cluster effectively over the ground truth classes when trained unsupervised, our GM-DGM can.
As our GM-DGM can also do semi-supervised learning at the same overall level of performance as SSVAEs, it can also capture semi-unsupervised learning.
As one might expect, the final performance in semi-unsupervised learning for GM-DGMs is above that for unsupervised learning and below vanilla semi-supervised learning.
Our GM-DGM approach has been able to learn successfully to classify and cluster for all ground truth classes of data in comparison to SSVAEs, for all data sets studied.

\section{Related Work}

Semi-unsupervised learning has similarities to some varieties of zero-shot learning (ZSL \cite{Weiss2016}, though in zero-shot learning one generally has access to auxiliary `attribute' information at training time, which we do not.
Semi-unsupervised learning also has links to transfer learning, particularly methods that attempt to discover new classes of data \cite{Hsu2018, Hsu2019, Han2019}.
That the assumptions of semi-supervised learning may not always hold is discussed in \cite{Oliver2018}.

Various other clustering and semi-supervised algorithms have the potential for use in semi-unsupervised learning.
The Cluster-aware Generative Model \cite{Maaloe2017} can, like our approach, learn in both unsupervised and semi-supervised regimes.
Generative Adversarial Networks \cite{Goodfellow2014} have also been used to approach semi-supervised learning and clustering.
Categorical Generative Adversarial Networks \cite{Springenberg2015} and Adversarial autoencoders \cite{Makhzani2016}, where the $\KL$ divergence in a VAE's ELBO is replaced with a GAN-like discriminator, can each learn in both regimes.

\section{Conclusion}
We introduced semi-unsupervised learning, an extreme limit case of semi-supervised learning where for some classes of data in our dataset there are no labelled examples at all.
We show that a common approach to semi-supervised learning using DGMs does not work for data of this type, even when making allowances for the nature of the training data.

To capture semi-unsupervised learning, we propose that a model must to be able to cluster data in a way that matches the structure given by the limited label information.
We call this the \textit{inductive bias requirement}.
And for the sake of ease of use, it is desirable to have an explicit representation of the posterior predictions of classes for any labelled data we have to train on.
Such an object can then be used directly as a classifier at test time.
We call this the \textit{classifier requirement}.
We have demonstrated that a simple Gaussian mixture deep generative model, with appropriate amortised variational posterior, can fulfill these requirements, learning successfully in the semi-unsupervised regime for the datasets studied.

We hope for further study of this new learning regime, as it is potentially the true state of affairs when learning with limited or biased labelled data.

\bibliographystyle{myIEEEtran}
\bibliography{references.bib}

\appendices

\newpage
\section{SSVAE Confusion Matrices}
\label{seq:M2confmat}

\begin{figure}[h!]
\centering
\makebox[\textwidth]{
\subfloat[SSVAE unsupervised on Fashion-MNIST\label{subfig:m2_us_fmnist_conf}]{%
      \includegraphics[width = 7.5cm]{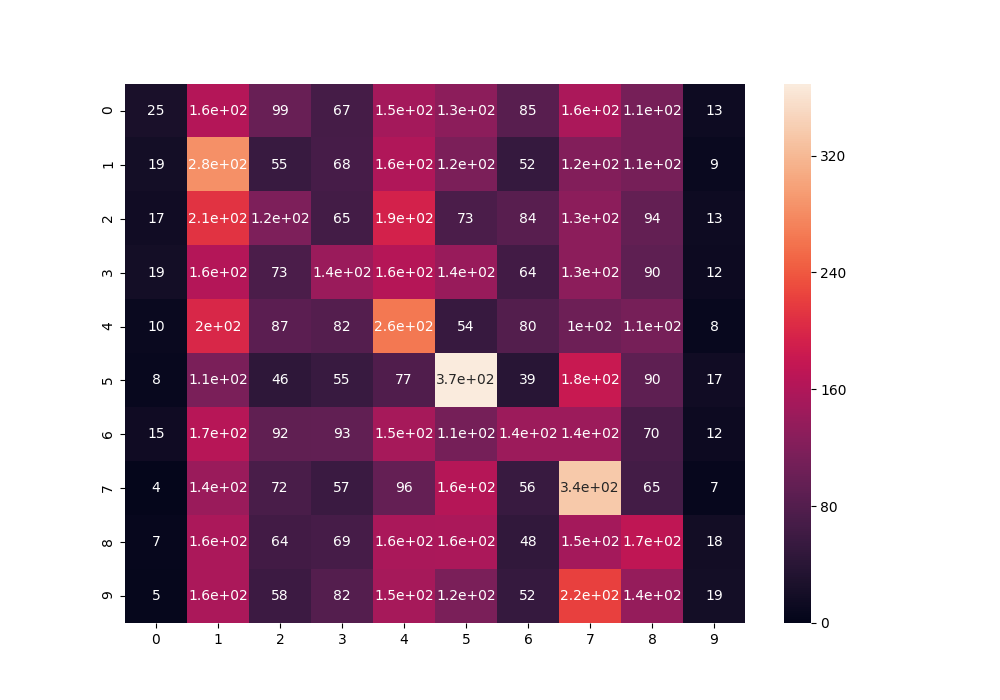}
     }
    }
\makebox[\textwidth]{%
\subfloat[SSVAE semi-supervised on Fashion-MNIST\label{subfig:m2_ss_fmnist_conf}]{%
      \includegraphics[width = 7.5cm]{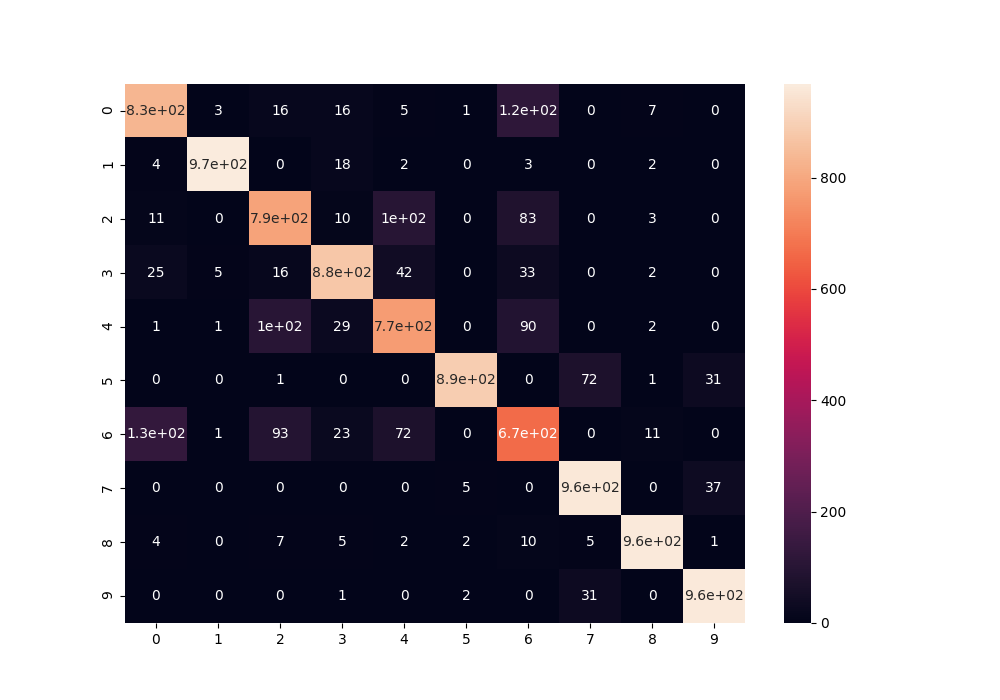}
     }  
     }
\makebox[\textwidth]{%
\subfloat[SSVAE semi-\textit{un}supervised on Fashion-MNIST\label{subfig:m2_sus_fmnist_conf}]{%
      \includegraphics[width = 7.5cm]{M2_plots/fmnist/m2_fmnist_sus_mlp_0_20_M2__fmnist__64__500__15__10_5_1_2_0015_0_conf.png}
     }  
}
     
\caption{Confusion matrix from Fashion-MNIST test set for SSVAEs. a) clearly shows that this model struggles to learn to partition the data into clusters corresponding to the ground truth classes. b) reiterates that these models do perform vanilla semi-supervised learning well. c) shows how on the unsupervised subproblem within semi-unsupervised learning this model also struggles. Recall that for c) classes 5-9 were entirely unlabelled in the training set.}
\label{fig:ssvae_conf}
\end{figure}

\newpage
\section{GM-DGM Confusion Matrices}
\label{seq:GMDGMconfmat}

\begin{figure}[h!]
\centering
\makebox[\textwidth]{%
\subfloat[GM-DGM unsupervised on Fashion-MNIST\label{subfig:gmdgm_us_fmnist_conf}]{%
      \includegraphics[width = 7.5cm]{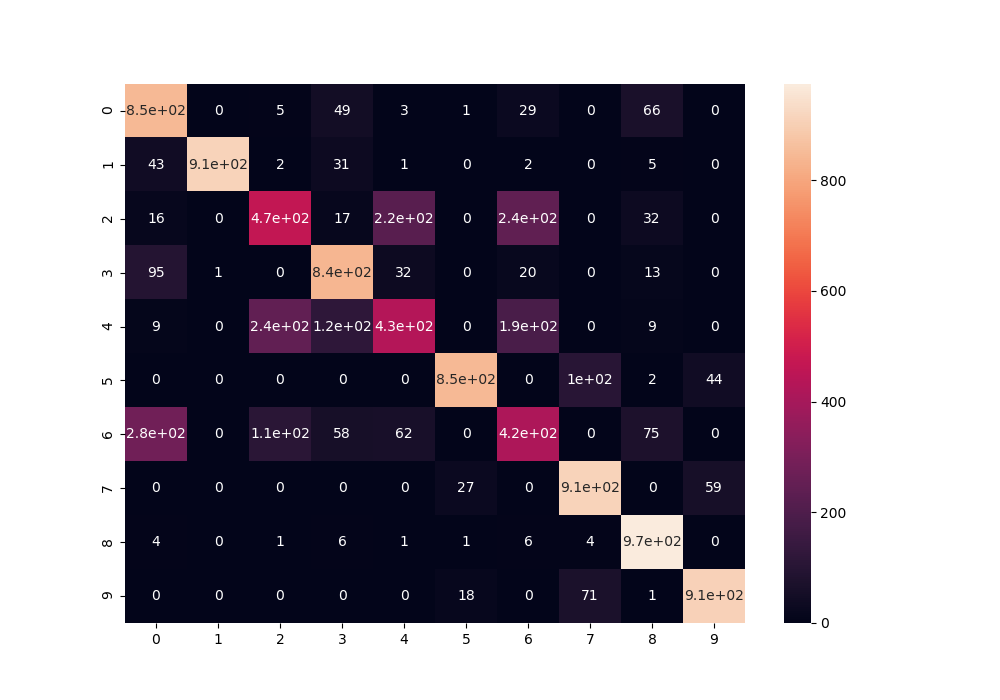}
     }
     }
\vspace{1mm}
\makebox[\textwidth]{%
\subfloat[GM-DGM semi-supervised on Fashion-MNIST\label{subfig:gmdgm_ss_fmnist_conf}]{%
      \includegraphics[width=7.5cm]{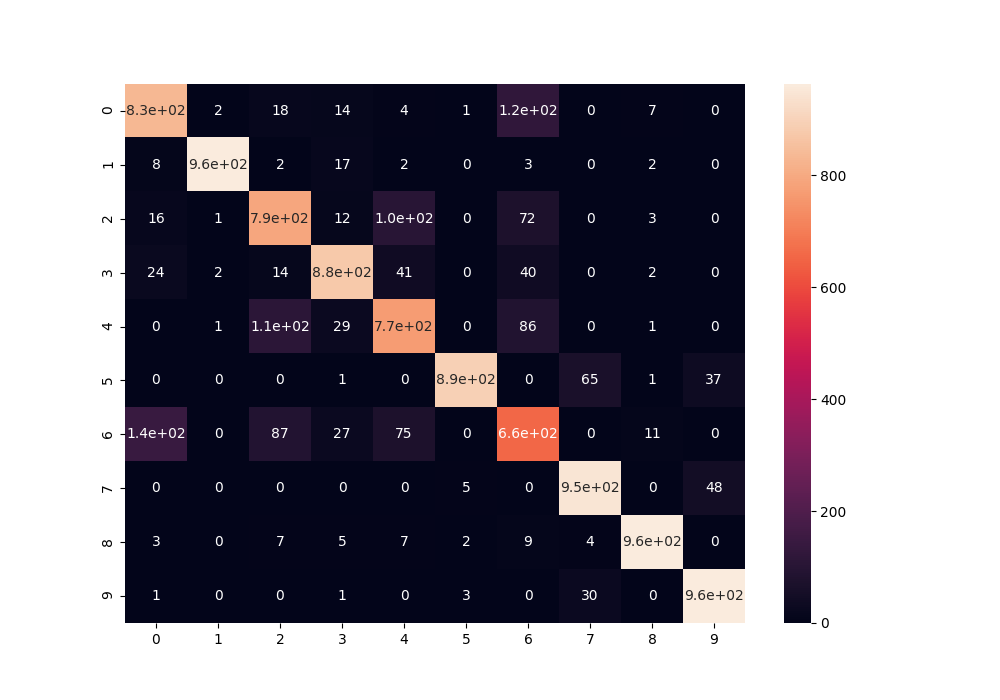}
     }
     }
    \vspace{1mm}
\makebox[\textwidth]{%
\subfloat[GM-DGM semi-\textit{un}supervised on Fashion-MNIST\label{subfig:gmdgm_sus_fmnist_conf}]{%
      \includegraphics[width=7.5cm]{GM_DGM_plots/fmnist/gm_dgm_fmnist_sus_mlp_0_20_GMDGM__fmnist__64__500__50__10_5_1_2_0015_0_conf.png}
     }  
    }
\caption{Confusion matrix for Fashion-MNIST test set from GM-DGMs. In all learning regimes, be it: a) unsupervised; b) semi-supervised; or c) semi-unsupervised this method is able to learn to separate the ground truth classes as well or better than SSVAEs.  Recall that for c) classes 5-9 were entirely unlabelled in the training set.}
\label{fig:gmdgm_conf}
\end{figure}

\newpage
\twocolumn
\section{Model Implementation and Data Preprocessing}
\label{ap:model}
\vspace{3mm}
The distributions $p_\theta(\v{x}|\cdot)$, $q_\phi(y|\v{x})$ and $q_\phi(\v{z}|\v{x},y)$ have their parameters parameterised by neural networks.
All networks are small MLPs.
$q_{\phi}(\v{z}|y,\v{x})$ in the variational posterior outputs the mean and (log) diagonal covariance for a Gaussian distribution.
Both in SSVAEs and GM-DGMs, the mean and variance networks have the same layers up to the second hidden layer, each having its own output layer.
For GM-DGMs, $p_\theta(\v{z}|y)$ is a look-up table: it maps a given setting of $y$ to its $\v{\mu}_\theta(y)$ and $\log \v{\Sigma}_\theta(y)$.
Between models, identical network architectures are used for components with the same purpose.
\begin{table}[H]
\caption{Per-dataset hyperparameters for networks in: $p_\theta(\v{x}|\cdot)$, $q_\phi(y|\v{x})$ and $q_\phi(\v{z}|\v{x},y)$.}
\begin{center}
\begin{small}
\begin{sc}
    \begin{tabular*}{\textwidth}{c @{\extracolsep{\fill}} cccc}
    \toprule
        Dataset & Dim $\v{z}$ & Units &  Batch Size & LR \\ \midrule
        MNIST        & 5  & 200                    & 4           & 0.001 \\
        Fashion-MNIST& 10 & 500                    & 64          & 0.0015\\
        HAR          & 15 & 500                    & 64          & 0.005 \\ \bottomrule
    \end{tabular*}
\end{sc}
\end{small}
\end{center}
    \label{tab:my_label}
\end{table}

Kernel initialisation is from a Gaussian distribution with standard deviation of $0.001$.
Biases are initialised with zeros.
We perform stochastic gradient ascent to maximise the ELBO in each case.
We use Adam \cite{Kingma2015}, with default moment parameters.
For the objectives of these models we must approximate various expectations wrt $q_\phi(\v{z}|\v{x},y)$ and $q_\phi(y|\v{x})$, which we do using the Reparameterisation Trick and Gumbel-Softmax Trick/Concrete Sampling respectively, taking 1 sample in each case per datapoint in the batch.
We trained for up to 400 epochs, with cosine decay of the learning rate.
For both MNIST and Fashion-MNIST we kept only dimensions of the data with a standard deviation greater than 0.1, and our likelihood function is a set of Bernoulli distributions.
We binarise the input image data during training, taking the greyscale values as the probabilities of pixels being set to one, taking a draw to represent each image in a batch. For HAR we used a Gaussian likelihood with fixed diagonal $\Sigma = \sigma^2 \mathbb{I}$ with $\sigma=0.01$.

\end{document}